\documentclass[conference]{IEEEtran}
\IEEEoverridecommandlockouts
\usepackage{comment}
\usepackage{cite}
\usepackage{amsmath,amssymb,amsfonts}
\usepackage{algorithm}
\usepackage{siunitx}
\usepackage{adjustbox}
\usepackage{multirow}
\usepackage{multicol}
\usepackage{placeins}

\usepackage{algpseudocode} 
\usepackage{graphicx}
\usepackage{textcomp}
\usepackage{xcolor}
\def\BibTeX{{\rm B\kern-.05em{\sc i\kern-.025em b}\kern-.08em
    T\kern-.1667em\lower.7ex\hbox{E}\kern-.125emX}}
\begin{document}

\title{TabFairGAN: Fair Tabular Data Generation with Generative Adversarial Networks}

\author{\IEEEauthorblockN{1\textsuperscript{st} Amirarsalan Rajabi}
\IEEEauthorblockA{\textit{Department of Computer Science} \\
\textit{University of Central Florida}\\
Orlando, FL, US \\
amirarsalan@knights.ucf.edu}
\and
\IEEEauthorblockN{2\textsuperscript{nd} Ozlem Ozmen Garibay}
\IEEEauthorblockA{\textit{Department of Industrial Engineering and Management Systems} \\
\textit{University of Central Florida}\\
Orlando, FL, US \\
ozlem@ucf.edu}
}

\maketitle

\begin{abstract}
With the increasing reliance on automated decision making, the issue of algorithmic fairness has gained increasing importance. In this paper, we propose a Generative Adversarial Network for tabular data generation. The model includes two phases of training. In the first phase, the model is trained to accurately generate synthetic data similar to the reference dataset. In the second phase we modify the value function to add fairness constraint, and continue training the network to generate data that is both accurate and fair. We test our results in both cases of unconstrained, and constrained fair data generation. In the unconstrained case, i.e. when the model is only trained in the first phase and is only meant to generate accurate data following the same joint probability distribution of the real data, the results show that the model beats state-of-the-art GANs proposed in the literature to produce synthetic tabular data. Also, in the constrained case in which the first phase of training is followed by the second phase, we train the network and test it on four datasets studied in the fairness literature and compare our results with another state-of-the-art pre-processing method, and present the promising results that it achieves. Comparing to other studies utilizing GANs for fair data generation, our model is comparably more stable by using only one critic, and also by avoiding major problems of original GAN model, such as mode-dropping and non-convergence, by implementing a Wasserstein GAN.

\end{abstract}


\section{Introduction}

Artificial intelligence has gained paramount importance in the contemporary human life. With an ever-growing body of research and increasing processing capacity of computers, machine learning systems are being adopted by many firms and institutions for decision-making. Various industries such as insurance companies, financial institutions, and healthcare providers rely on automated decision making by machine learning models, making fairness-aware learning crucial since many of these automated decisions could have major impacts on the lives of individuals.

There are numerous evidence suggesting that bias exists in AI systems. One well known example is Correctional Offender Management Profiling for Alternative Sanctions (COMPAS), which is a decision making system deployed by the US criminal justice system to assess the likelihood of a criminal defendant's recidivism (re-offending). It is shown that COMPAS is biased against African American defendants \cite{chouldechova2017fair}. Another example is a Google's targeted advertising that was found to have shown the high paid jobs significantly more to males than females \cite{lambrecht2019algorithmic}.

The existence of such bias and unfair classifications in AI systems has led the research community to pay attention to the problem of bias in AI. There are different approaches to improve fairness existing in the AI fairness literature. Let $D=\{X,S,Y\}$ be a labelled dataset, where $X\in\mathbb{R}^n$ are the unprotected attributes, $S$ is the protected attribute, and $Y$ is the decision. From a legal perspective, protected attribute is the attribute identified by law, based on which it is illegal to discriminate \cite{pessach2020algorithmic}, e.g. gender or race. The proposed fairness enforcement methods in the literature could be categorised into three main classes of \emph{pre-process} methods, \emph{in-process} methods, and \emph{post-process} methods. 

Pre-process methods include modifying the training data before feeding the data into machine learning algorithm. For instance, in one study \cite{kamiran2012data}, four methods are presented to remove bias including suppression which is to remove attributes highly correlated with protected attributes $S$, massaging the dataset which is to change labels ($Y$) of some objects in the dataset, and reweighing that involves assigning weights to different instances in the dataset. These are preliminary and simpler methods that results in more fair predictions, however entail higher fairness-utility cost. In other words fairness is achieved at the expense of accuracy. Another preprocessing method proposed in the literature is the work of Feldman et al. \cite{feldman2015certifying} in which a repairment mechanism is proposed to modify the unprotected attributes ($X$) and achieve fairness with higher accuracy comparing to the aforementioned methods. This method will be discussed in more detail in Section~\ref{certifying} as the baseline method. In-process approaches involve modifying the learning algorithm to achieve fairness during training \cite{pessach2020algorithmic}. These methods mostly include modifying the objective functions or adding regularization terms to the cost function. For example, \cite{kamishima2012fairness} proposes adding a regularization term to the objective function which penalize mutual information between the protected attributes and the classifier predictions. Finally, post-process mechanisms include modifying the final decisions of the classifiers. For instance, Hardt et al. \cite{hardt2016equality} propose a method to modify the final classification scores in order to enhance equalized odds.

The emergence of unfairness in AI systems is mostly attributed to: 1) direct bias existing in the historical datasets being used to train the algorithms, 2) bias caused by missing data, 3) bias caused by proxy attributes, where bias against the minority population is present in non-protected attributes, and 4) bias resulting from algorithm objective functions, where the aggregate accuracy of the whole population is sought and therefore the algorithm might disregard the minority group for the sake of majority \cite{pessach2020algorithmic}. Since historical datasets are a major source of discrimination in AI, we focus on generating unbiased datasets to achieve fairness.

There is a rich and growing literature on generative models. The main idea behind a generative model is to capture the probabilistic distribution that could generate data similar to a reference dataset \cite{oussidi2018deep}. Broadly speaking, generative models could be divided into two main classes of models \cite{oussidi2018deep}: Energy-based models such as Boltzmann Machines \cite{fahlman1983massively}, and  cost function-based models such as autoencoders and generative adversarial networks (GANs) \cite{goodfellow2014generative}. GANs address some deficiencies in traditional generative models, and are shown to excel in various tasks comparing to other generative models such as in image generation \cite{brock2018large} and video generation \cite{vondrick2016generating}. 

The original GAN consists of two networks, \emph{generator} and \emph{discriminator} \cite{goodfellow2014generative}. The two networks play a \emph{minimax} game. The generator takes a latent random variable $Z$ as input and generates a sample $G(Z)$, that is similar to the real data. The discriminator, on the other hand, is fed with both real and generated samples, and its task is to correctly classify the input sample as real or generated. Over time if the networks have enough capacity, they are trained together and ideally optimized to reach an equilibrium state in which the generator produces data from the exact targeted distribution and the discriminator gives the real and generated samples an equal probability of 0.5. The work in \cite{goodfellow2014generative} shows that training the discriminator to optimality is equal to minimizing Jensen-Shannon divergence \cite{menendez1997jensen}. The work of Arjovsky et al. develops Wasserstein GANs, where a \emph{critic} replaces the discriminator, and minimizing Earth-mover's distance is used instead of minimizing Jensen-Shannon divergence \cite{rubner2000earth}. They show that WGAN could address some common training problems attributed to GANs, usch as requirement to maintain a careful balance during training as well as mode dropping \cite{arjovsky2017wasserstein}.

In recent studies adversarial training has been used to remove discrimination. One such study, for example, by formulating the model as a minimax problem, proposes an adversarial learning framework that could learn representations of data that are discrimination-free and do not contain explicit information about the protected attribute \cite{edwards2015censoring}. Other adversarial objectives are proposed by the works of \cite{madras2018learning, zhang2018mitigating} to achieve group fairness measures such as demographic parity and equality of odds. The application of generative adversarial networks for fairness in tabular datasets is not discussed enough in the literature, but has recently attracted attention of the research community. For instance, the work of Sattigeri et al. \cite{sattigeri2019fairness} proposes an approach to generate image datasets such that \emph{demographic fairness} in the generated dataset is imposed. In their work Xu et al. \cite{xu2018fairgan} design a GAN that produces discrimination free tabular datasets. Their network includes one generator and two discriminators. The generator is adopted from \cite{choi2017generating} and produces fake pairs of data $(\hat{X}, \hat{Y})$ following the conditional distribution $P_G(X,Y|S)$. One discriminator's task is to ensure generator produces data with good accuracy, and the second discriminator ensures the generator produces fair data.

In this paper, we propose a Wasserstein GAN, TabFairGAN, that can produce high quality tabular data with the same joint distribution as the original tabular dataset. In Section~\ref{sec:dp}, we discuss the fairness measure: demographic parity and discrimination score. In Section~\ref{model-description}, we introduce the model architecture, data transformation, value functions, and the training process of the model. In section~\ref{datagen_nofairness}, we compare the results of TabFairGAN with two other state-of-the-art GANs for tabular data generation, namely TGAN \cite{xu2018synthesizing} and CTGAN \cite{xu2019modeling}. In Section\ref{datagen_withfairness}, we show how the model could be used for fair data generation and test the model on four real dataset. We compare the results of our model with the method developed by \cite{feldman2015certifying}, which is another pre-process methods to enforce fairness. Finally in Section~\ref{fairness_tradeoff}, we explore the fairness-accuracy trade-off. This work has two main contributions. We show that in the case of no constraints present (no fairness), the model is able to produce high quality synthetic data, competing with the state-of-the-art GANs designed for tabular data generation. Second contribution is producing high quality fair synthetic data, by adding a fairness constraint in the loss function of the generator. Comparing our model to previous application of GANs for fair tabular data generation, the model is more stable based on two merits: 1) the proposed model is a Wasserstein GAN which is shown to improve original GAN model in terms of some common GAN pitfalls, such as mode-dropping phenomena \cite{arjovsky2017wasserstein}, and 2) the model only uses one critic instead of two \cite{xu2018fairgan} or three \cite{xu2019fairgan+} discriminators.

\section{Discrimination Score}
\label{sec:dp}
Among the most frequently practiced fairness metrics specified in legal notions and the literature is \emph{demographic parity} or statistical parity/fairness. The goal of demographic fairness is to ensure that the overall proportion of members with respect to the protected group receiving a positive decision is identical. In a binary case, let $D=\{X,S,Y\}$ be a labelled dataset, where $X\in\mathbb{R}^n$ is the unprotected attributes, $S$ is the protected attribute, and $Y$ is the decision. In this paper, we consider the binary case, and for notational convenience we assume that the protected attribute $S$ takes two values, where $S=0$ represents the underprivileged minority class, and $S=1$ represents the privileged majority class. For instance, in a binary racial discrimination study the value $0$ will be assigned to ``African-American'', whereas $1$ is assigned to ``White''. We also assign $1$ to $Y$ for a successful decision (for instance an admission to a higher education institution), and assign $0$ to $Y$ for an unsuccessful decision (rejection). Demographic fairness for the labeled dataset is defined as follows \cite{hardt2016equality}:

\begin{equation}
    P(y=1|s=1) = P(y=1|s=0)
\end{equation}

In this context, demographic parity is defined by the difference between the conditional probability and its marginal. We define the discrimination with respect to the protected attribute $S$ by \emph{discrimination score} (DS) and calculate it by: $DS = P(y=1|s=1) - P(y=1|s=0)$. A similar measure could be obtained for a labeled dataset $D$ and a classifier $f : (X,S) \rightarrow Y$ where the discrimination score for the classifier $f$ with respect to protected attribute $S$ can be obtained by:

\begin{equation}
    P(\hat{y}=1|x,s=1) - P(\hat{y}=1|x,s=0)
\end{equation}

\section{Model Description}
\label{model-description}
\subsection{Tabular Dataset Representation and Transformation}
A tabular dataset contains $N_C$ numerical columns $\{c_{1},...,c_{N_C}\}$ and $N_D$ categorical columns $\{d_{1},...,d_{N_D}\}$. In this model, categorical columns are transformed and represented by one-hot vectors. Representing numerical columns on the other hand is non-trivial due to certain properties of numerical columns. One such property is that numerical columns are often sampled from multi-modal distributions. Some models such as \cite{choi2017generating} use min-max normalization to normalize and transform numerical columns. The work of Xu et al. \cite{xu2019modeling} proposes a more complex process, namely a mode-specific normalization using variational Gaussian mixture model (VGM) to estimate the number of modes and fit a Gaussian mixture model to each numerical column. In our model, each numerical column is transformed using a quantile transformation \cite{beasley2009rank}:
\begin{equation}
    c^{'}_{i} = \Phi^{-1}(F(c_{i}))
\end{equation}

Where $c_{i}$ is the $i$th numerical feature, $F$ is the CDF (cumulative distbituion function) of the feature $c_{i}$, and $\Phi$ is the CDF of a uniform distribution. After transforming numerical and discrete columns, the representation of each transformed row of the data is as follows:
\begin{align}
    &\textbf{r} = c^{'}_{1} \oplus ... \oplus c^{'}_{N_C} \oplus d^{'}_{1} \oplus ... \oplus d^{'}_{N_D} \\
    &l_i = dim(d^{'}_{i}) \\
    &l_w = dim(r)
\end{align}

where $c{'}_{i}$ represents the $i$th numerical column, $d{'}_{i}$ denotes the one-hot encoded vector of the $i$th categorical columns, and $\oplus$ is the symbol denoting concatenation of vectors. Also, $l_i$ shows the dimension of the $i$th discrete column's one-hot encoding vector and $l_w$ shows the the dimension of $r$.

\subsection{Network Structure}
While traditional GANs suffer from problems such as non-convergence and mode-collapse, the work of \cite{arjovsky2017wasserstein} developed Wasserstein GANs which improve training of GANs to some extent, and replace the discriminator with a \emph{critic}. The network designed in this model is a WGAN with gradient penalty \cite{NIPS2017_892c3b1c}. The WGAN value function using the Kantorovich-Rubinstein duality \cite{villani2009optimal} is as follows \cite{NIPS2017_892c3b1c}:

\begin{equation}
    \min_{G}\max_{C \in \mathcal{C}}\mathop{\mathbb{E}}_{\mathbf{x} \sim P_{data}(\mathbf{x})}[C(\mathbf{x})]- \mathop{\mathbb{E}}_{\mathbf{z} \sim P_{z}(\mathbf{z})}[C(G(z))]
\end{equation}

Where $\mathcal{C}$ is the set of 1-Lipschitz functions. The generator receives a latent variable $Z$ from a standard multivariate normal distribution and produces a sample data point which is then forwarded to the critic. Once the critic and the generator are trained together, eventually the generator would become like a deterministic transformation that produces data similar to the real data.

The generator consists of a fully-connected first layer with ReLu activation function. The second hidden layer of the generator network is then formed by concatenation of multiple vectors that could form data similar to transformed original data. For the numerical variables, a fully connected layer of $\text{FC}_{l_w \rightarrow N_C}$, with a ReLu activation is implemented. For nodes that are supposed to produce discrete columns, multiple fully connected layer of $\text{FC}_{l_w \rightarrow l_i}$, with Gumble softmax \cite{jang2016categorical} activation is used in order to produce one-hot vectors ($d^{'}_{i}$). The resulting nodes are then concatenated to produce data similar to the transformed original data (with the same dimension of $l_w$), which is then fed to the critic network. The structure of the critic network is simple and includes 2 fully connected layers with Leaky ReLu activation functions. 

The generator network's architecture is formally described as:

\begin{equation}
  \begin{cases}
    h_{0} = Z \text{ (latent vector)} \\
    h_{1} = \text{ReLu}(\text{FC}_{l_w \rightarrow l_w}(h_0)) \\
    h_{2} = \text{ReLu}(\text{FC}_{l_w \rightarrow N_C}(h_1)) \oplus \text{gumble}_{0.2}(\text{FC}_{l_w \rightarrow l_1}(h_1)) \oplus \\ \text{gumble}_{0.2}(\text{FC}_{l_w \rightarrow l_2}(h_1)) \oplus \text{...} \oplus \text{gumble}_{0.2}(\text{FC}_{l_w \rightarrow l_{N_D}}(h_1))
    
  \end{cases}
\end{equation}

Where $FC_{a \rightarrow b}$ denotes a fully connected layer with input size $a$ and output size $b$, $\text{ReLu}(x)$ shows applying a ReLu activation on $x$, and $\text{gumble}_{\tau}(x)$ denotes applying Gumble softmax with parameter $\tau$ on a vector $x$, and $\oplus$ denotes concatenation of vectors. 

The critic network's architecture is formally described as:

\begin{equation}
  \begin{cases}
    h_{0} =  X \text{ (output of the generator or transformed real data)} \\
    h_{1} = \text{LeakyReLu}_{0.01}(\text{FC}_{l_w \rightarrow l_w}(h_0)) \\
    h_{2} = \text{LeakyReLu}_{0.01}(\text{FC}_{l_w \rightarrow l_w}(h_1))
  \end{cases}
\end{equation}

\begin{figure}
	\includegraphics[width=0.98\linewidth,keepaspectratio]{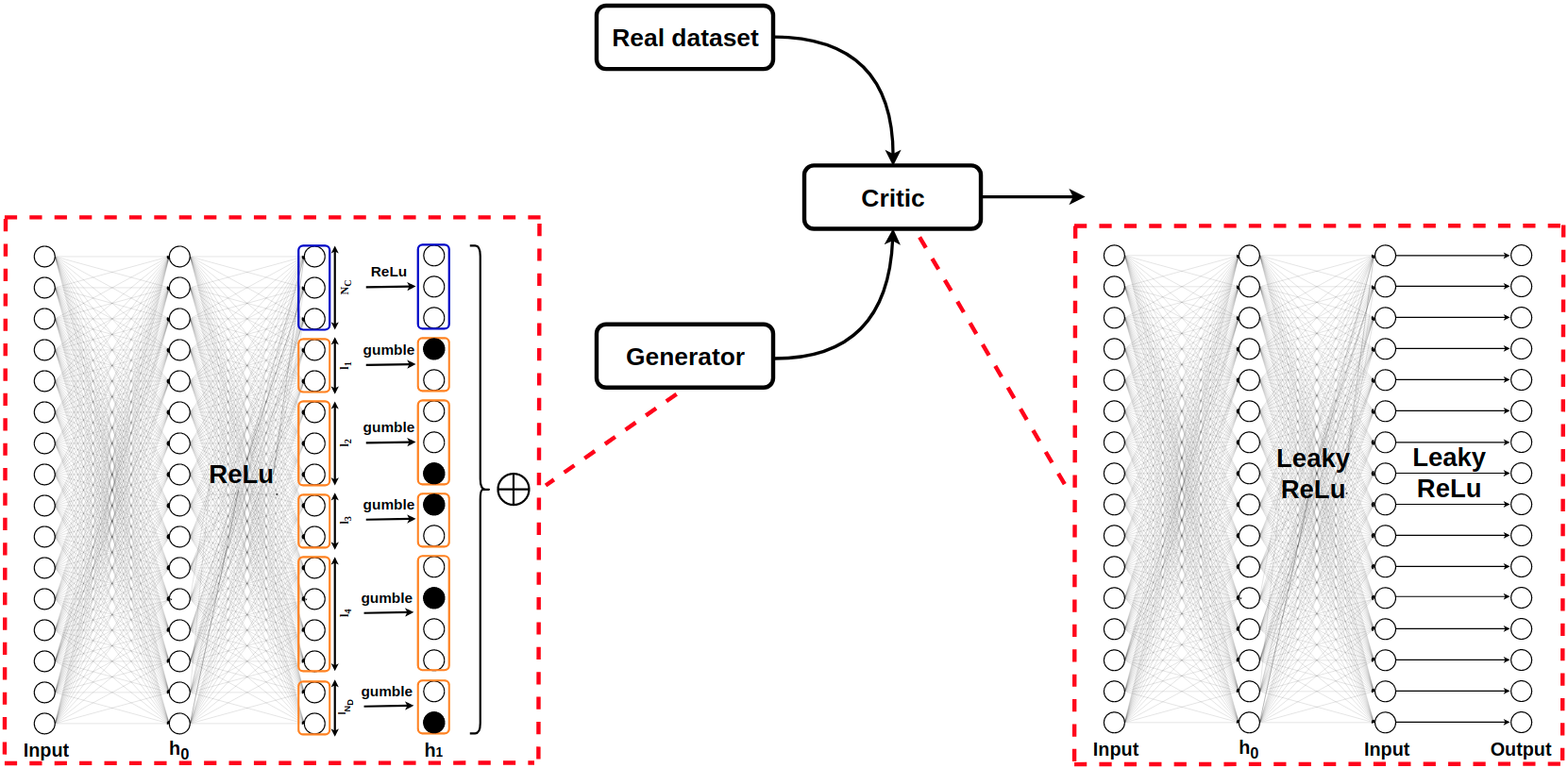}
	\caption{Model architecture. The generator consists of an initial fully connected layer with ReLu activation function, and a second layer which uses ReLu for numerical attributes generation and gumble-softmax to form one-hot representations of categorical attributes. The final data is then produced by concatenating all attributes in the last layer of the generator. The critic consists of fully-connected layers with LeakyReLu activation function.}
	\label{fig-modelarch}
\end{figure}

Where $\text{LeakyReLu}_{\tau}(x)$ denotes applying Leaky ReLu activation function \cite{xu2015empirical} with slope $\tau$ on $x$. Fig.~\ref{fig-modelarch} shows the architecture of the model.

\subsection{Training}
\label{subsec_training}
In this section we introduce the loss functions for the critic network and generator network of the developed WGAN. The overall process of training the model includes two phases. \textit{Phase I} of training only focuses on training the model such that the generator could generate data with a joint probability distribution similar to that of the real data. \textit{Phase II} of training further trains the generator to produce samples which have a joint probability distribution similar to that of real data and is also fair, with respect to discrimination score (DS) defined in Section~\ref{sec:dp}.

\subsubsection{Phase I: Training for Accuracy}
In the first phase, generator and critic are trained with respect to their value functions. Critic's loss function with gradient penalty is \cite{NIPS2017_892c3b1c}:

\begin{equation}
    V_C = \mathop{\mathbb{E}}_{\mathbf{\hat{x}} \sim P_{g}}[C(\hat{\mathbf{x}})] - \mathop{\mathbb{E}}_{\mathbf{x} \sim P_r}[C(\mathbf{x})] + \lambda \mathop{\mathbb{E}}_{\mathbf{\bar{x}} \sim P_{\bar{\mathbf{x}}}} [(||\nabla_{\bar{\mathbf{x}}}C(\bar{\mathbf{x}})||_{2}-1)^2]
\end{equation}

Where $P_{\mathbf{r}}$ and $P_{\mathbf{g}}$ are real data distribution and generated data distribution, respectively. Note that the third term is the gradient penalty to enforce the Lipschitz constraint, and $P_{\mathbf{\bar{x}}}$ is implicitly defined sampling uniformly along straight lines between pairs of points sampled from the data distribution $P_{\mathbf{r}}$ and the generator distribution $P_{\mathbf{g}}$ \cite{NIPS2017_892c3b1c}. 

The loss function for the generator network in Phase I of training is also as follows:

\begin{equation}
    V_G = -\mathop{\mathbb{E}}_{\mathbf{\hat{x}} \sim P_{g}}[C(\hat{\mathbf{x}})]
\end{equation}

\subsubsection{Phase II: Training for Fairness and Accuracy}
In the second phase of training, fairness constraint is enforced on generator to produce fair data. Similar to the definitions in Section~\ref{sec:dp}, let $\hat{D} = \{\hat{X}, \hat{Y}, \hat{S}\}$ be a batch of generated data, where $\hat{X}$ is the unprotected attribute of the generated data, $\hat{Y}$ is the decision with $\hat{Y}=1$ being the successful and favorable value for the decision (e.g. having an income of $>50K$ for an adult in the adult income dataset), and $\hat{S}$ being the protected attribute with $\hat{S}=0$ showing the unprivileged minority group (for example having a gender of ``female'' in the adult income data set). The new loss function for the generator in Phase II of training is as follows:
\begin{equation}
\begin{split}
    V_G = -&\mathop{\mathbb{E}}_{(\mathbf{\hat{x},\hat{y},\hat{s})} \sim P_{g}}[C(\mathbf{\hat{x}, \hat{y}, \hat{s}})] - \lambda_f ( \mathop{\mathbb{E}}_{(\mathbf{\hat{x},\hat{y},\hat{s})} \sim P_{g}} [\mathbf{\hat{y}}|\mathbf{\hat{s}}=0] - \\ &\mathop{\mathbb{E}}_{(\mathbf{\hat{x},\hat{y},\hat{s})} \sim P_{g}} [\mathbf{\hat{y}}|\mathbf{\hat{s}}=1])
\end{split}
\end{equation}
With the above loss function for the generator, the model aims to generate a fair dataset $\{\hat{X}, \hat{Y}, \hat{S}\} \sim P_{g}$ which achieves the demographic fairness with respect to the protected attribute $\hat{S}$ in the generated samples, by minimizing discrimination score in the generated data $P(\hat{Y}|\hat{S}=1)-P(\hat{Y}|\hat{S}=0)$. The goal in this phase of training is to train the generator to generate synthetic data which is both similar to the real data $\hat{D} \sim D$, and the generated data is fair based on demographic fairness measure. In the ideal case, the generator would produce synthetic data such that $\hat{Y} \perp \hat{S}$. After training is done, the samples are generated and inverse transformed to the original data format. The formal procedure of training the model is shown in Algorithm~\ref{algo}.

\begin{algorithm}
\caption{training algorithm for the proposed WGAN. We use $n_{crit}=4$, batch size of 256, $\lambda_p=10$, Adam optimizer with $\alpha=0.0002$, $\beta_1=0.5$, and $\beta_2=0.999$}
	\begin{algorithmic}[1]
		\For {$T_1$}
			\For {$t=1,\ldots,n_{crit}$}
				\State \begin{scriptsize} Sample batch $m$ $D(x,y,s) \sim P_r$ and $z \sim P(z)$ and $\epsilon \sim U[0,1]$
				\end{scriptsize}
				\State $\hat{D} = (\hat{x},\hat{s},\hat{y}) \leftarrow G_{\theta}(z)$
				\State $\bar{D} \leftarrow \epsilon(D)+(1-\epsilon)(\hat{D})$
				\State Update the critic by descending the gradient:
				
				\State \begin{footnotesize}
				$\nabla_{w} \frac{1}{m} \sum_{i=1}^{m}C_w(\hat{D})-C_w(D) + \lambda_{p}(||\nabla_{\bar{D}}C_{w}(\bar{D})||_{2}-1)^2$               
				\end{footnotesize}
		        
			\EndFor
			\State Sample a batch $m$ $z \sim P(z)$
			\State Update the generator by descending the gradient:
			\State $\nabla_{\theta} \frac{1}{m}\sum_{i=1}^{m}-(C_{w}(G_{\theta}(z)))$
		\EndFor
		
		\For {$T_2$}
		    \For {$t=1,\ldots,n_{crit}$}
				\State \begin{scriptsize} Sample batch $m$ $D(x,y,s) \sim P_r$ and $z \sim P(z)$ and $\epsilon \sim U[0,1]$
				\end{scriptsize}
				\State $\hat{D} = (\hat{x},\hat{s},\hat{y}) \leftarrow G_{\theta}(z)$
				\State $\bar{D} \leftarrow \epsilon(D)+(1-\epsilon)(\hat{D})$
				\State Update the critic by descending the gradient:
		        \State \begin{footnotesize} $\nabla_{w} \frac{1}{m} \sum_{i=1}^{m}C_w(\hat{D})-C_w(D) + \lambda_{p}(||\nabla_{\bar{D}}C_{w}(\bar{D})||_{2}-1)^2$
		        \end{footnotesize}
			\EndFor
		    \State sample a batch $m$ $\hat{D}={\hat{x},\hat{s},\hat{y}} \sim P(G_{\theta}(z))$
		    \State Update the generator by descending the gradient:
		    \State \begin{small} $\nabla_{\theta} \frac{1}{m} \sum_{i=1}^{m}-C_{w}(\hat{D})-\lambda_{f}(\frac{|D_{s=0,y=1}|}{|D_{s=0}|}-\frac{|D_{s=1,y=1}|}{|D_{s=1}|})$
		    \end{small}
		\EndFor
	\end{algorithmic}
	\label{algo}
\end{algorithm}

\section{Experiment: only Phase I (no fairness)}
\label{datagen_nofairness}
In this section, we evaluate the effectiveness of the model in producing synthetic data simialr to data coming from a known probability distribution. We show that the model is able to generate synthetic data similar to the reference dataset and compare our results with two state-of-the-art GAN models for generation of tabular datasets, namely TGAN \cite{xu2018synthesizing} and CTGAN \cite{xu2019modeling}. TGAN is a GAN-based model that generates relational tables by clustering numerical variables to deal with multi-modal distributions and adding noise and KL divergence into loss function to generate discrete features. In CTGAN, mode-specific normalization is applied to numerical values and the generator works conditionally in order to overcome the imbalance in training data. We evaluate the model on UCI Adult Income Dataset\footnote{http://archive.ics.uci.edu/ml/datasets/adult} \cite{Dua:2019}.
The task we are trying to achieve is as follows: given a dataset $D=\{X,S,Y\} \sim P_{data}$, generate a dataset $\hat{D}_{syn}=\{\hat{X}, \hat{S}, \hat{Y}\} \sim P_{syn}$ s.t. $P_{syn} \sim P_{data}$. We are not seeking to achieve fairness in this section, and we solely seek to generate data following the same distribution as real data to achieve data utility.

To compare data utility among generated datasets among different models, we evaluate the performance of using synthetic data as training data for machine learning. At first, the real dataset is divided into two parts: $\mathbf{D_{train}}$ and $\mathbf{D_{test}}$. Adult dataset contains a total of 48,842 rows. 90\% of the data were assigned to $\mathbf{D_{train}}$ and the rest 10\% were assigned to $\mathbf{D_{test}}$. Next, each model is trained on the training set $\mathbf{D_{train}}$ for 300 epochs three times. With each training, the trained models are used to generate their corresponding synthetic data $\mathbf{D_{syn}}$. Three machine learning classifiers are then chosen and trained on each generated $\mathbf{D_{syn}}$, tested on $\mathbf{D_{test}}$, and eventually the accuracy and F1 score of classification is recorded. The classifiers used are a Decision Tree Classifier, Logistic Regression, and a Multi Layer Perceptron. Table~\ref{generation_comp} reports the results of classification, and compares the results with the case that a classifier is trained on the original $\mathbf{D_{train}}$, and tested on $\mathbf{D_{test}}$ (reporting the means and standard deviations of evaluation metrics).
The results shows that TabFairGAN and CTGAN outperform TGAN in all cases. TabFairGAN outperforms CTGAN with a DT Classifier. With a LR classifier, the performance of TabFairGAN and CTGAN is identical with respect to accuracy, and TabFairGAN performs slightly better than CTGAN with respect to F1 score. With a MLP classifier, CTGAN performs slightly better than TabFairGAN with respect to accuracy, while TabFairGAN outperforms CTGAN with respect to F1 score. These results display the effetiveness of TabFariGAN with respect to generating data identical to real tabular data.

\begin{table*}
\caption{Comparing the results TabFairGAN for accurate data generation with TGAN and CTGAN models}
\label{generation_comp}
\centering
\def\arraystretch{1.5}
\begin{tabular}{|c|l|l|l|l|l|l|}
\hline
\multirow{2}{*}{Classifier} & \multicolumn{2}{|c|}{DTC}                                         & \multicolumn{2}{c|}{LR}                                         & \multicolumn{2}{c|}{MLP}                                          \\ 
\cline{2-7}
                            & \multicolumn{1}{c|}{Accuracy}   & \multicolumn{1}{c|}{F1}         & \multicolumn{1}{c|}{Accuracy}   & \multicolumn{1}{c|}{F1}        & \multicolumn{1}{c|}{Accuracy}   & \multicolumn{1}{c|}{F1}          \\ \hline
Original Data               &$ 0.811\pm0.001$                   &$0.606\pm0.002$                   &$0.798\pm0.000$                   &$0.378\pm0.000$                  & $0.780\pm0.051$                   & $0.488\pm0.075$                    \\ \hline
TabFairGan                  & $\textbf{0.783}\pm\textbf{0.001}$ & $\textbf{0.544}\pm\textbf{0.002}$ & $\textbf{0.794}\pm\textbf{0.020}$ & $\textbf{0.239}\pm\textbf{0.012}$ & $0.778\pm0.045$                   & $\textbf{0.405}\pm\textbf{0.174}$  \\ \hline
TGAN                        & $0.661\pm0.013$                   & $0.503\pm0.012$                   & $0.765\pm0.010$                   & $0.170\pm0.008$                  & $0.623\pm0.197$                   & $0.376\pm0.159$                    \\ \hline
CTGAN                       & $0.777\pm0.003$                   & $0.482\pm0.004$                   & $\textbf{0.794}\pm\textbf{0.023}$          & $0.232\pm0.012$                  & $\textbf{0.784}\pm\textbf{0.007}$ & $0.305\pm0.104$ \\ \hline                   
\end{tabular}
\end{table*}

\section{Experiments: Fair Data Generation and Data Utility (training with both Phase I and Phase II)}
\label{datagen_withfairness}
In the second set of experiments, the effectiveness of the model in generating data which is both similar to the reference dataset and also fair is evaluated, and the tradeoff between machine learning efficacy and fairness is investigated. We will experiment with four datasets to test the fairness/utility tradeoff of the model. The four datasets and their attributes are first introduced. All four datasets used in experiments are studied in the literature of algorithmic fairness \cite{pessach2020algorithmic}. Next, we introduce the baseline method with which the results of TabFairGAN are compared. The results are presented and compared in Table~\ref{results_table}.

\subsection{Datasets}

The first dataset is UCI Adult Dataset \cite{Dua:2019}. This dataset is based on 1994 US census data and contains 48,842 rows with attributes such as age, sex, occupation, and education level. for each person, and the target variable indicates whether that individual has an income that exceeds \$50K per year or not. In our experiments, we consider the protected attribute to be sex ($S=\text{``Sex''}$, $Y=\text{``Income''}$). 

The second dataset used in the experiments is the Bank Marketing Data Set \cite{moro2014data}. This dataset contains information about a direct marketing campaign of a Portuguese banking institution. Each row of the dataset contains attributes about an individual such as age, job, marital status, housing, duration of that call, and the target variable determines whether that individual subscribed a term deposit or not. The dataset contains 45,211 records. Similar to \cite{zafar2017fairness}, we have considered age to be the protected attribute (a young individual has a higher chance of being labeled as ``yes'' to subscribe a term deposit). In order to have a binary protected attribute, we set a cut-off value of 25 and an age of more than 25 is considered ``older'', while an age of less than or equal to 25 is considered ``younger'' ($S=\text{``Age''}$, $Y=\text{``Subscribed''}$). 

The third dataset used in this section is the ProPublica dataset from COMPAS risk assessment system \cite{ProPublica}. This dataset contains information about defendants from Broward County, and contains attributes about defendants such as their ethnicity, language, marital status, sex, etc. ,and for each individual a score showing the likelihood of recidivism (re-offending). In this experiments we used a modified version of the dataset. First, attributes such as FirstName, LastName, MiddleName, CASE\_ID, and DateOfBirth are removed. Studies have shown that this dataset is biased against African Americans \cite{chouldechova2017fair}. Therefore, ethnicity is chosen to be the protected attribute for this study. Only African American and Caucasian individuals are kept and the rest are dropped. The target variable in this dataset is a risk decile score provided by COMPAS system, showing the likelihood of that individual to re-offend, which ranges from 1 to 10. The final modified dataset contains 16,267 records with 16 features. To make the target variable binary, a cut-off value of 5 is considered and individuals with a declile score of less than 5 are considered ``Low\_Chance'', while the rest are considered ``High\_Chance''. ($S=\text{``Ethnicity''}$, $Y=\text{``Recidivism\_Chance''}$). 

The last dataset used in experiments is the Law School Admission Council which is made by conducting a survey across 162 law schools in the United States \cite{wightman1998lsac}. This dataset contains information on 21,790 law students such as their GPA (grade-point average), LSAT score, race, and the target variable is whether the student had a high FYA (first year average grade). Similar to other studies (such as \cite{bechavod2017penalizing}), we have considered race to be the protected attribute. We only considered individuals with ``Black'' or ``White'' race. The modified data contains 19,567 records. ($S=\text{``Race''}$, $Y=\text{``FYA''}$). There discrimination score (DS) of all datasets are reported in Table~\ref{results_table}.

\subsection{Baseline Model: Certifying and Removing Disparate Impact}\label{certifying}
In their work Feldman et al. \cite{feldman2015certifying} proposed a method to modify a dataset to remove bias and preserve relevant information in the data. In dataset $D=\{X,S,Y\}$, given the protected attribute $S$  and a \emph{single} numerical attribute $X$, let $X_s=Pr(X|S=s)$ denote the marginal distribution on $X$ conditioned on $S=s$. Considering $F_s:X_s \rightarrow [0,1]$ the cumulative distribution function for values $x \in X_s$, they define a ``median'' distribution $A$ in terms of its quantile function $F^{-1}_{A}: F^{-1}_{A}(u)=\text{median}_{s \in S}F^{-1}_{s}(u)$. They then propose a repair algorithm which creates $\bar{X}$, such that for all $x \in X_s$ the corresponding $\bar{x} = F^{-1}_{A}(F_{s}(x))$. To control the trade-off between fairness and accuracy, they define and calculate \emph{$\lambda-\text{partial repair}$} by:

\begin{equation}
    \bar{F}^{-1}_{s} = (1-\lambda)F^{-1}_{s} + \lambda(F_A)^{-1}
\end{equation}

The result of such partial repair procedure is a dataset $\bar{D}=\{\bar{X},S,Y\}$ which is more fair and preserves relevant information for classification task. We call this method \emph{CRDI} henceforth.

\subsection{Results}

The goal in this section is to train the proposed network on datasets and produce similar data that is also fair with respect to protected attributes defined for each dataset. The process is as follows: The models are first trained on each dataset. As mentioned in Section~\ref{subsec_training}, training of the network includes two phases: in the first phase, the network is only trained for accuracy for a certain number of epochs, and then in the second phase, the loss function of generator is modified and the network gets trained for accuracy and fairness. Once the training is finished, the generator of the network is used to produce synthetic data $\mathbf{D_{syn}}$. We also generated repaired datasets using CRDI method described in Section~\ref{certifying} to compare our results with. For each model, we train five times and report the means and standard deviations of evaluation results in Table~\ref{results_table}. 

The generated data $\mathbf{D_{syn}}$ is then evaluated from two perspective: fairness and utility. To evaluate the fairness of $\mathbf{D_{syn}}$, we adopt discrimination score (DS): $DS = P(y=1|s=1) - P(y=1|s=0)$. Looking into Table~\ref{results_table}, the results show that comparing with CRDI, TabFairGAN could more effectively produce datasets s.t. demographic parity in the generated data is almost removed. The demographic parity of the produced datasets by TabFairGAN, beat the repaired datasets produced by CRDI. 

To evaluate data utility, we adopt a decision tree classifier with the default parameter setting \cite{pedregosa2011scikit}. For TabFairGAN data, We train the decision tree classifier on $\mathbf{D_{syn}}$ and test it on $\mathbf{D_{test}}$, and report the accuracy and F1-score of the classifier. We also train decision tree classifiers on repaired data $\bar{D}$ produced by CRDI, and test on $\mathbf{D_{test}}$ and report accuracy and f1-score. Table~\ref{results_table} shows that repaired data $\bar{D}$ produced by CRDI has better data utility for adult dataset, COMPAS dataset, and Law School dataset by less than 5\% in all cases, while the accuracy of $\mathbf{D_{syn}}$ produced by TabFairGAN is almost 8\% higher than that of $\bar{D}$ produced by CRDI. 

The last evaluation we perform on the produced datasets is to examine discrimination score (DS) of the classifier. we adopt discrimination score (DS) for classifier: $DS = P(\hat{y}=1|s=1) - P(\hat{y}=1|s=0)$. The results in Table~\ref{results_table} show that discrimination score of the decision tree classifier trained on  $\mathbf{D_{syn}}$ for Adult dataset and Law School is lower by almost 4\% and 13\%, respectively, while the discrimination score of the decision tree classifier trained on  $\mathbf{\bar{D}}$ for Bank dataset and COMPAS dataset is lower by 1\% and 0.003\%, respectively.

The parameter settings of the models on each datasets is reported in the Appendix. The results show, while CRDI narrowly beats TabFairGAN in terms of data utility, TabFairGAN beats CRDI in terms of discrimination score in all cases for generated data and in 2 out of 4 cases in the generated classifiers. This is attributed to fairness utility trade-off of TabFairGAN governed by $\lambda_f$. The case of COMPAS dataset is interesting since none of the models could decrease discrimination score in the classifier much, comparing to the discrimination score in the original dataset. Looking into the data and performing a correlation analysis, risk decile score (target variable) has a high Pearson correlation of 0.757 with one of columns names RecSupervisionLevel which denotes the supervisory status of each individual. This reveals that although the generated dataset $\mathbf{D_{syn}}$ has a lower discrimination score of 0.009, disparate impact exists in the dataset, indicating that the discriminatory outcomes are not explicitly caused by the protected attribute, but are also from the proxy unprotected attributes \cite{xu2018fairgan}.

\begin{table*}[h]
\caption{Comparing the results of TabFairGAN for fair data generation with CRDI}
\label{results_table}
\centering
\begin{adjustbox}{width=0.96\textwidth}
\def\arraystretch{1.5}

\setlength\extrarowheight{3pt}
\Large

\begin{tabular}{|c|c|c|c|c|c|c|c|c|c|c|c|}
\hline
\multicolumn{1}{|l}{} & \multicolumn{3}{|c|}{\textbf{Original Data}}              & \multicolumn{4}{c|}{\textbf{TabFairGAN}}         & \multicolumn{4}{c|}{\textbf{CRDI}}        \\
\cline{1-12}
\textbf{Dataset}              & \textbf{Orig. Acc.}   & \textbf{F1 Orig.}     & \textbf{DS in Orig. Data}  & \textbf{DS Gen. Data}  & \textbf{Acc. Gen. Data}                 & \textbf{F1 Gen. Data}                    & \textbf{DS in Classifier}          & \textbf{DS Rep. Data.}   & \textbf{Acc. Rep. Data}                  & \textbf{F1 Rep. Data}                    & \textbf{DS in Classifier}             \\ \hline
\textbf{Adult}               & $0.816\pm0.005$ & $0.619\pm0.013$ & 0.195          & $\textbf{0.009}\pm\textbf{0.027}$ & $0.773\pm0.013$ & $0.536\pm0.022$  & $\textbf{0.082}\pm\textbf{0.038}$      & $0.165\pm 0.048$ & $\textbf{0.793}\pm\textbf{ 0.011}$ & $\textbf{0.558}\pm\textbf{ 0.029}$ & $0.121\pm 0.024$   \\ \hline
\textbf{Bank}                 & $0.879\pm0.004$ & $0.491\pm0.020$ & $0.126$          & $\textbf{0.001}\pm\textbf{0.011}$  & $\textbf{0.854}\pm\textbf{0.004}$ & $0.373\pm0.024$  & $0.060\pm0.056$                 & $0.122\pm0.004$  & $0.776\pm0.004$   & $\textbf{0.384}\pm\textbf{0.011}$  & $\textbf{0.050}\pm\textbf{0.017}$   \\ \hline
\textbf{COMPAS}               & $0.903\pm0.007$ & $0.914\pm0.007$ & 0.258         & $\textbf{0.009}\pm\textbf{0.102}$  & $0.860\pm0.040$ & $0.876\pm0.033$  & $0.208\pm0.072$                  & $0.119\pm0.128$ & $\textbf{0.893}\pm\textbf{0.021}$  & $\textbf{0.906}\pm\textbf{0.020}$                     & $\textbf{0.205}\pm\textbf{0.055}$  \\ \hline
\textbf{Law School}           & $0.854\pm0.008$ & $0.918\pm0.005$ & 0.302         & $\textbf{0.024}\pm\textbf{0.036}$  & $0.847\pm0.020$ & $0.916\pm0.012$  & $\textbf{0.153}\pm\textbf{0.072}$       & $0.233\pm0.103$ & $\textbf{0.892}\pm\textbf{0.004}$  & $\textbf{0.941}\pm\textbf{0.002}$  & $0.289\pm0.057$  \\ \hline      
\end{tabular}
\end{adjustbox}
\end{table*}

\subsection{Utility and Fairness Trade-off}
\label{fairness_tradeoff}
\begin{figure}
	\includegraphics[width=0.90\linewidth,keepaspectratio]{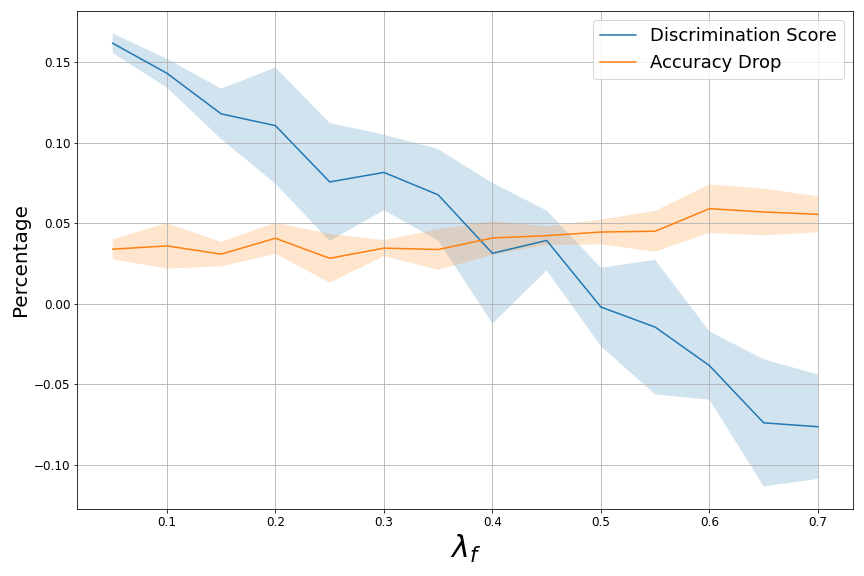}
	\caption{Exploring the trade-off between accuracy and fairness by incremental increasing of parameter $\lambda_{f}$}
	\label{fig-lambda}
\end{figure}

To explore the trade-off between utility and fairness of the generated data, we perform the following experiment: $\lambda_f$ was increased between $[0.05, 0.7]$ in steps of 0.05, and for each value of $\lambda_f$ the model was trained 170 epochs in phase I and 30 times in the phase II. For each $\lambda_f$ value, five training was performed and the average of Discrimination Score was recorded for each $\lambda_f$. Figure~\ref{fig-lambda} shows the results, plotted along with standard deviation as confidence intervals. We can observe that discrimination score of the generated synthetic datasets ($D_{syn}$) is decreasing significantly as $\lambda_f$ decreases. Meanwhile, classifier accuracy layoff, i.e. the reduction in decision tree classifier's accuracy comparing to the case in which the classifier is trained on the real original training dataset ($D_{train}$), is increasing slightly as $\lambda_f$ increases.

\section{Conclusion}
In this paper, we proposed a Wasserstein Generative Adversarial Network that could generate synthetic data similar to a reference data. We showed that in the case of unconditional tabular data generation, i.e. with no fairness constrains, the model is able to produce data with high quality comparing to other GANs developed for the same purpose. We also showed that by adding a fairness constraint to the generator, the model is able to achieve data generation which improves the demographic parity of the generated data. We tested the model on four datasets studies in the fairness literature and compared our results with that of \cite{feldman2015certifying}. As a generative model, GANs have a great potential to be utilized for fair data generation, specially in the case that the real dataset is limited. There are other field in which GANs could be utilized for tabular data generation, such as the research involved with data privacy \cite{park2018data}. In the future work, we will explore other more sophisticated data generation constraints, e.g. considering enforcing other fairness metrics such as equality of odds and equality of opportunity. We also consider exploring utilizing GANs for fairness in other data types, such as text and image data.

\bibliographystyle{IEEEtran}
\bibliography{IEEEabrv,mybib}

\begin{thebibliography}{10}
\providecommand{\url}[1]{#1}
\csname url@samestyle\endcsname
\providecommand{\newblock}{\relax}
\providecommand{\bibinfo}[2]{#2}
\providecommand{\BIBentrySTDinterwordspacing}{\spaceskip=0pt\relax}
\providecommand{\BIBentryALTinterwordstretchfactor}{4}
\providecommand{\BIBentryALTinterwordspacing}{\spaceskip=\fontdimen2\font plus
\BIBentryALTinterwordstretchfactor\fontdimen3\font minus
  \fontdimen4\font\relax}
\providecommand{\BIBforeignlanguage}[2]{{%
\expandafter\ifx\csname l@#1\endcsname\relax
\typeout{** WARNING: IEEEtran.bst: No hyphenation pattern has been}%
\typeout{** loaded for the language `#1'. Using the pattern for}%
\typeout{** the default language instead.}%
\else
\language=\csname l@#1\endcsname
\fi
#2}}
\providecommand{\BIBdecl}{\relax}
\BIBdecl

\bibitem{chouldechova2017fair}
A.~Chouldechova, ``Fair prediction with disparate impact: A study of bias in
  recidivism prediction instruments,'' \emph{Big data}, vol.~5, no.~2, pp.
  153--163, 2017.

\bibitem{lambrecht2019algorithmic}
A.~Lambrecht and C.~Tucker, ``Algorithmic bias? an empirical study of apparent
  gender-based discrimination in the display of stem career ads,''
  \emph{Management Science}, vol.~65, no.~7, pp. 2966--2981, 2019.

\bibitem{pessach2020algorithmic}
D.~Pessach and E.~Shmueli, ``Algorithmic fairness,'' \emph{arXiv preprint
  arXiv:2001.09784}, 2020.

\bibitem{kamiran2012data}
F.~Kamiran and T.~Calders, ``Data preprocessing techniques for classification
  without discrimination,'' \emph{Knowledge and Information Systems}, vol.~33,
  no.~1, pp. 1--33, 2012.

\bibitem{feldman2015certifying}
M.~Feldman, S.~A. Friedler, J.~Moeller, C.~Scheidegger, and
  S.~Venkatasubramanian, ``Certifying and removing disparate impact,'' in
  \emph{proceedings of the 21th ACM SIGKDD international conference on
  knowledge discovery and data mining}, 2015, pp. 259--268.

\bibitem{kamishima2012fairness}
T.~Kamishima, S.~Akaho, H.~Asoh, and J.~Sakuma, ``Fairness-aware classifier
  with prejudice remover regularizer,'' in \emph{Joint European Conference on
  Machine Learning and Knowledge Discovery in Databases}.\hskip 1em plus 0.5em
  minus 0.4em\relax Springer, 2012, pp. 35--50.

\bibitem{hardt2016equality}
M.~Hardt, E.~Price, and N.~Srebro, ``Equality of opportunity in supervised
  learning,'' \emph{Advances in neural information processing systems},
  vol.~29, pp. 3315--3323, 2016.

\bibitem{oussidi2018deep}
A.~Oussidi and A.~Elhassouny, ``Deep generative models: Survey,'' in \emph{2018
  International Conference on Intelligent Systems and Computer Vision
  (ISCV)}.\hskip 1em plus 0.5em minus 0.4em\relax IEEE, 2018, pp. 1--8.

\bibitem{fahlman1983massively}
S.~E. Fahlman, G.~E. Hinton, and T.~J. Sejnowski, ``Massively parallel
  architectures for al: Netl, thistle, and boltzmann machines,'' in
  \emph{National Conference on Artificial Intelligence, AAAI}, 1983.

\bibitem{goodfellow2014generative}
I.~Goodfellow, J.~Pouget-Abadie, M.~Mirza, B.~Xu, D.~Warde-Farley, S.~Ozair,
  A.~Courville, and Y.~Bengio, ``Generative adversarial nets,'' \emph{Advances
  in neural information processing systems}, vol.~27, 2014.

\bibitem{brock2018large}
A.~Brock, J.~Donahue, and K.~Simonyan, ``Large scale gan training for high
  fidelity natural image synthesis,'' \emph{arXiv preprint arXiv:1809.11096},
  2018.

\bibitem{vondrick2016generating}
C.~Vondrick, H.~Pirsiavash, and A.~Torralba, ``Generating videos with scene
  dynamics,'' \emph{Advances in neural information processing systems},
  vol.~29, pp. 613--621, 2016.

\bibitem{menendez1997jensen}
M.~Men{\'e}ndez, J.~Pardo, L.~Pardo, and M.~Pardo, ``The jensen-shannon
  divergence,'' \emph{Journal of the Franklin Institute}, vol. 334, no.~2, pp.
  307--318, 1997.

\bibitem{rubner2000earth}
Y.~Rubner, C.~Tomasi, and L.~J. Guibas, ``The earth mover's distance as a
  metric for image retrieval,'' \emph{International journal of computer
  vision}, vol.~40, no.~2, pp. 99--121, 2000.

\bibitem{arjovsky2017wasserstein}
M.~Arjovsky, S.~Chintala, and L.~Bottou, ``Wasserstein generative adversarial
  networks,'' in \emph{International conference on machine learning}.\hskip 1em
  plus 0.5em minus 0.4em\relax PMLR, 2017, pp. 214--223.

\bibitem{edwards2015censoring}
H.~Edwards and A.~Storkey, ``Censoring representations with an adversary,''
  \emph{arXiv preprint arXiv:1511.05897}, 2015.

\bibitem{madras2018learning}
D.~Madras, E.~Creager, T.~Pitassi, and R.~Zemel, ``Learning adversarially fair
  and transferable representations,'' in \emph{International Conference on
  Machine Learning}.\hskip 1em plus 0.5em minus 0.4em\relax PMLR, 2018, pp.
  3384--3393.

\bibitem{zhang2018mitigating}
B.~H. Zhang, B.~Lemoine, and M.~Mitchell, ``Mitigating unwanted biases with
  adversarial learning,'' in \emph{Proceedings of the 2018 AAAI/ACM Conference
  on AI, Ethics, and Society}, 2018, pp. 335--340.

\bibitem{sattigeri2019fairness}
P.~Sattigeri, S.~C. Hoffman, V.~Chenthamarakshan, and K.~R. Varshney,
  ``Fairness gan: Generating datasets with fairness properties using a
  generative adversarial network,'' \emph{IBM Journal of Research and
  Development}, vol.~63, no. 4/5, pp. 3--1, 2019.

\bibitem{xu2018fairgan}
D.~Xu, S.~Yuan, L.~Zhang, and X.~Wu, ``Fairgan: Fairness-aware generative
  adversarial networks,'' in \emph{2018 IEEE International Conference on Big
  Data (Big Data)}.\hskip 1em plus 0.5em minus 0.4em\relax IEEE, 2018, pp.
  570--575.

\bibitem{choi2017generating}
E.~Choi, S.~Biswal, B.~Malin, J.~Duke, W.~F. Stewart, and J.~Sun, ``Generating
  multi-label discrete patient records using generative adversarial networks,''
  in \emph{Machine learning for healthcare conference}.\hskip 1em plus 0.5em
  minus 0.4em\relax PMLR, 2017, pp. 286--305.

\bibitem{xu2018synthesizing}
L.~Xu and K.~Veeramachaneni, ``Synthesizing tabular data using generative
  adversarial networks,'' \emph{arXiv preprint arXiv:1811.11264}, 2018.

\bibitem{xu2019modeling}
L.~Xu, M.~Skoularidou, A.~Cuesta-Infante, and K.~Veeramachaneni, ``Modeling
  tabular data using conditional gan,'' in \emph{Advances in Neural Information
  Processing Systems}, 2019.

\bibitem{xu2019fairgan+}
D.~Xu, S.~Yuan, L.~Zhang, and X.~Wu, ``Fairgan+: Achieving fair data generation
  and classification through generative adversarial nets,'' in \emph{2019 IEEE
  International Conference on Big Data (Big Data)}.\hskip 1em plus 0.5em minus
  0.4em\relax IEEE, 2019, pp. 1401--1406.

\bibitem{beasley2009rank}
T.~M. Beasley, S.~Erickson, and D.~B. Allison, ``Rank-based inverse normal
  transformations are increasingly used, but are they merited?'' \emph{Behavior
  genetics}, vol.~39, no.~5, pp. 580--595, 2009.

\bibitem{NIPS2017_892c3b1c}
\BIBentryALTinterwordspacing
I.~Gulrajani, F.~Ahmed, M.~Arjovsky, V.~Dumoulin, and A.~C. Courville,
  ``Improved training of wasserstein gans,'' in \emph{Advances in Neural
  Information Processing Systems}, I.~Guyon, U.~V. Luxburg, S.~Bengio,
  H.~Wallach, R.~Fergus, S.~Vishwanathan, and R.~Garnett, Eds., vol.~30.\hskip
  1em plus 0.5em minus 0.4em\relax Curran Associates, Inc., 2017. [Online].
  Available:
  \url{https://proceedings.neurips.cc/paper/2017/file/892c3b1c6dccd52936e27cbd0ff683d6-Paper.pdf}
\BIBentrySTDinterwordspacing

\bibitem{villani2009optimal}
C.~Villani, \emph{Optimal transport: old and new}.\hskip 1em plus 0.5em minus
  0.4em\relax Springer, 2009, vol. 338.

\bibitem{jang2016categorical}
E.~Jang, S.~Gu, and B.~Poole, ``Categorical reparameterization with
  gumbel-softmax,'' \emph{arXiv preprint arXiv:1611.01144}, 2016.

\bibitem{xu2015empirical}
B.~Xu, N.~Wang, T.~Chen, and M.~Li, ``Empirical evaluation of rectified
  activations in convolutional network,'' \emph{arXiv preprint
  arXiv:1505.00853}, 2015.

\bibitem{Dua:2019}
\BIBentryALTinterwordspacing
D.~Dua and C.~Graff, ``{UCI} machine learning repository,'' 2017. [Online].
  Available: \url{http://archive.ics.uci.edu/ml}
\BIBentrySTDinterwordspacing

\bibitem{moro2014data}
S.~Moro, P.~Cortez, and P.~Rita, ``A data-driven approach to predict the
  success of bank telemarketing,'' \emph{Decision Support Systems}, vol.~62,
  pp. 22--31, 2014.

\bibitem{zafar2017fairness}
M.~B. Zafar, I.~Valera, M.~G. Rogriguez, and K.~P. Gummadi, ``Fairness
  constraints: Mechanisms for fair classification,'' in \emph{Artificial
  Intelligence and Statistics}.\hskip 1em plus 0.5em minus 0.4em\relax PMLR,
  2017, pp. 962--970.

\bibitem{ProPublica}
\BIBentryALTinterwordspacing
J.~Angwin, J.~Larson, S.~Mattu, and L.~Kirchner. (2016) Machine bias
  propublica. [Online]. Available:
  \url{https://www.propublica.org/article/machine-bias-risk-assessments-in-criminal-sentencing}
\BIBentrySTDinterwordspacing

\bibitem{wightman1998lsac}
L.~F. Wightman, ``Lsac national longitudinal bar passage study. lsac research
  report series.'' 1998.

\bibitem{bechavod2017penalizing}
Y.~Bechavod and K.~Ligett, ``Penalizing unfairness in binary classification,''
  \emph{arXiv preprint arXiv:1707.00044}, 2017.

\bibitem{pedregosa2011scikit}
F.~Pedregosa, G.~Varoquaux, A.~Gramfort, V.~Michel, B.~Thirion, O.~Grisel,
  M.~Blondel, P.~Prettenhofer, R.~Weiss, V.~Dubourg \emph{et~al.},
  ``Scikit-learn: Machine learning in python,'' \emph{the Journal of machine
  Learning research}, vol.~12, pp. 2825--2830, 2011.

\bibitem{park2018data}
N.~Park, M.~Mohammadi, K.~Gorde, S.~Jajodia, H.~Park, and Y.~Kim, ``Data
  synthesis based on generative adversarial networks,'' \emph{arXiv preprint
  arXiv:1806.03384}, 2018.

\end{thebibliography}

\section{Appendix}
Table~\ref{model_parameters} reports the models' hyperparameters used in Section~\ref{datagen_withfairness} experiments.

\begin{table}[h]
\caption{Model Parameters}
\label{model_parameters}
\centering
\def\arraystretch{1.5}
\begin{tabular}{|l|l|l|l|l|}
\hline
           & \multicolumn{3}{l|}{TabFairGAN} & CRDI    \\ \hline
           & T\_1 & T\_2 & Lambda           & Lambda  \\ \hline
Adult      & 170  & 30   & 0.5              & 0.999   \\ \hline
Bank       & 195  & 5    & 0.75             & 0.9     \\ \hline
COMPAS     & 40   & 30   & 2.2              & 0.999   \\ \hline
Law School & 180  & 20   & 2.5              & 0.999   \\ \hline
\end{tabular}
\end{table}

\end{document}